\documentclass[11pt,letterpaper]{llncs}
\usepackage{amsmath}
\usepackage{amssymb}
\usepackage{graphicx}
\usepackage{marvosym}
\usepackage{url}
\usepackage{fancyhdr}

\usepackage{geometry}
\newcommand{\keywords}[1]{\par\addvspace\baselineskip
\noindent\keywordname\enspace\ignorespaces#1}

\fancypagestyle{firstpage}{\fancyhf{}
\setcounter{page}{1}
\rfoot{\thepage}
}

\begin{document}


\title{\LARGE{Solar Potential Analysis of Rooftops Using Satellite Imagery}}


%
%
\author{\large{Akash Kumar}}
\institute{\large{Electronics and Communication Engineering, Delhi Technological University, \\ New Delhi-110042, India}}

%


%
%


\maketitle

\thispagestyle{firstpage}

\begin{abstract}
Solar energy is one of the most important sources of renewable energy and the cleanest form of energy. In India, where solar energy could produce power around trillion kilowatt-hours in a year, our country is only able to produce power of around in gigawatts only. Many people are not aware of the solar potential of their rooftop, and hence they always think that installing solar panels is very much expensive. In this work, we introduce an approach through which we can generate a report remotely that provides the amount of solar potential of a building using only its latitude and longitude. We further evaluated various types of rooftops to make our solution more robust. We also provide an approximate area of rooftop that can be used for solar panels placement and a visual analysis of how solar panels can be placed to maximize the output of solar power at a location.
\end{abstract}


\section{Introduction}

Solar power is the process of conversion of sunlight energy to electrical energy. Photovoltaic cells present on the solar panels convert the sunlight energy to Direct current, and then an inverter converts it to Alternative current source which gets supplied for household activities. Solar energy has several benefits ranging from habitat conservation to combat climate change. It is also the cheapest and reliable form of energy. Once installed, the solar panels can efficiently perform up to 20-25 years with minimal maintenance. However, in India, people hesitate to invest in solar panels as the procedure is very troublesome. The company sends their personnel, and they take measurement of each sides and obstacles present on the rooftop. After that, they build a report accordingly, which takes a lot of time. Sometimes, these personnel have to take multiple visits too. It takes a lot amount of time and wastage of money too. It is also impossible for an industry to individually analyze an area by surveying the solar potential of every house. Technology like Image Processing and Computer Vision plays a pivotal role to overcome this challenge. 

Many works have been done in foreign countries to automate the solar power analysis of an area. In the USA, Project Sunroof has been launched by Google to overcome this problem. They have used 3D mapping technology along with a high-quality satellite map that is available for the USA. Still, their technology is limited to forty-two cities of USA only. In contrast to this, none of the service providers has 3D maps available for India. In India, we have various types of map service providers. We explored Open Street Maps, MapMyIndia and Google Maps. After examining all of them, Google maps performed best in our case. Therefore, there is a dire need to create a more generalized solution that can adapt well to many situations.

\begin{figure*}[htbp]
\centering
\includegraphics[width=0.8\linewidth]{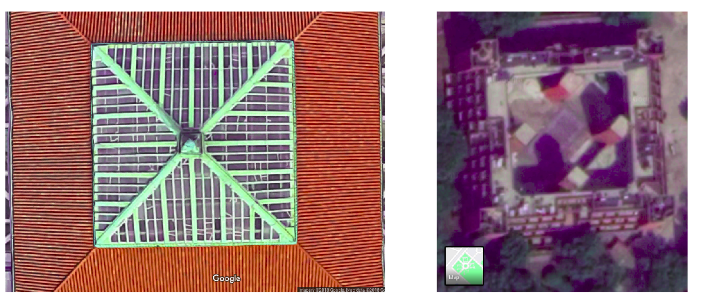}
\caption{Example of rooftop images from Google Maps. Left Image depicts the USA rooftop at zoom level 21 and right image depicts Indian building rooftop at zoom level 20. The rooftop can be clearly marked out in the case of the USA whereas in India it's not possible to visually mark out the rooftop boundaries (reproduced from https://www.google.com/maps)} \label{fig1}
\end{figure*}

Solar panels placement in case of India is very different as contrasted to the USA. In the USA, Google Satellite Maps can be zoomed up to level 21, and they also have a 3D mapping that helps to localize rooftops as well as obstacles(e.g., Water tanks and AC Inverter units) whereas in India the maps can be zoomed up to level 20 only with no 3D mapping. Example of Google Satellite map at the highest zoom level in the USA and India is shown in Fig.\ref{fig1}. From the figure, we can inspect the quality of maps in the USA as compared to India. While we can visually mark out the optimal area for solar panel placement in the USA, in India, we can't even crop out the exact rooftop at the highest zoom level. Therefore, our solution concentrates on detection of the usable rooftop area of a building, as well as, to maximize the solar power generation by orienting solar panels at a perfect angle.

In this paper, we employ several algorithms to address the problem of rooftop detection and optimal area localization. We created a dataset of 50 images on which exhaustive experimentation were done. The main contributions are:
\begin{enumerate}
    \item We proposed an approach to segment out optimal rooftop area for were  selected.  The  problem  with  corner  features  are  that solar panel placement in India where the image quality of satellite imagery is very deficient.
    
    \item We considered several types of the rooftop to learn the intra-class variations and to evaluate our method. Our final output provides a comparative analysis of the area that can account for solar panel placement. Our solution also layouts at which angles solar panels should be placed for maximum solar output.
\end{enumerate} 

The rest of the paper is organized as follows. In Section II, we walk through the brief overview of the related work. Section III presents the description of the dataset and Section IV describes the proposed approach in detail. Section V presents the experimentation and results analysis, followed by the Conclusion and Future Work in Section VI.

\section{Related Work}

There are several recent papers that have addressed the problems of identification of building rooftops \cite{ref_1}, \cite{ref_2}, \cite{ref_3}, \cite{ref_5}, most of them are based on 3D reconstruction, which uses Fast Graph search \cite{ref_1}, GIS maps\cite{ref_2} and other topology based automatic modeling. \cite{ref_3} identified the rooftops using a single remote sensing image compared to a stereo-based approach. Markov Random Field and projective geometry were used to detect rooftops accurately. 

Work has been done using LiDAR technology using Object-based \cite{ref_4} method to localize rooftops over an area. LiDAR technology automatically extracted building footprints and segment out the rooftop planes, but this technology is costly, and it's not affordable for start-ups to use this technology. Our main aim is to develop a solution that requires minimal investment to get the analysis of rooftops. Automatic building detection was procured using Spectral graph theory. It's suitable for the USA dataset only \cite{ref_22}.

Moreover, the houses in the USA have slanted rooftops whereas in India the rooftops are planes. There are works done on building detection and its 3D modeling in the USA. In India, there are no 3D maps available. These methods have massive shortcomings when applied.  The  problem  with  corner  features  are  that to Indian style rooftops.  Our method automatically detects the optimal area in case of Indian rooftops. The above methods failed miserably in case of planar rooftops. We explored the area of obstacle identification in the rooftop area using Satellite imagery and thus marked out the area available for solar power generation.

\begin{figure}[b]
\begin{center}
\includegraphics[scale=0.35]{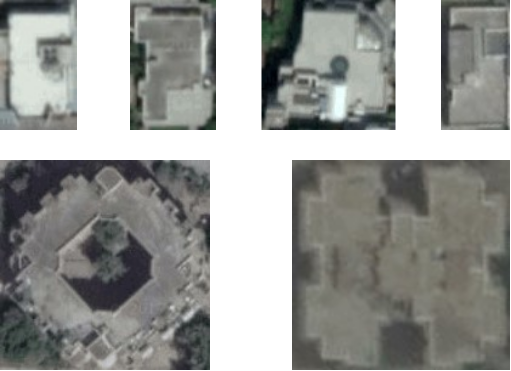}
\caption{Dataset Overview} \label{fig2}
\end{center}
\end{figure}

\section{Dataset}
The manually collected Indian aerial rooftop dataset consists of all types of variants. We considered all the circumstances of the types of rooftop obstacles and also multiple rooftop sizes. Currently, we have tested our algorithm on 50 different types of rooftops. Overview of the dataset is depicted in the diagram Fig.\ref{fig2}. 

\section{Proposed Approaches}
In this section, we discuss the proposed framework that is used for optimal rooftop detection from the dataset shown above. The devised architecture used to solve the problem is described as follows: 

\subsection{Aerial Rooftop Detection}

\subsubsection{Watershed Segmentation} This algorithm is mainly applied when we want to segment out objects-of-interest that are close to each other, especially in the cases where objects are placed very close to each other. It is based on the marker-based algorithm and uses distance transform. It treats the image as if it is a topological surface which has some high altitude parts and some low altitude parts. The brightness of the image is compared with this high and low peaks and valleys. High brightness portion is treated as peak and low brightness portion as a valley. This segmentation algorithm is mainly applied to grayscale images. 

\begin{figure*}[t]
\centering
\includegraphics[width=\textwidth]{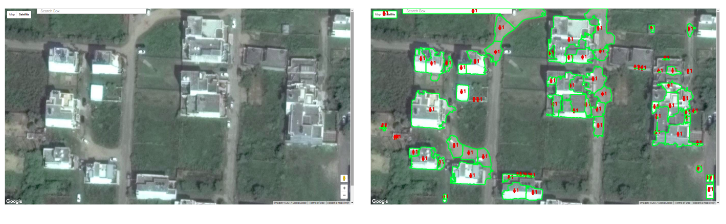}
\caption{Watershed Segmentation algorithm on Google Satellite Image on the left. On the right side, the output shows detection of buildings. It can be clearly seen that it is marking out roads also as buildings.} \label{fig3}
\end{figure*}

After a process of erosion and dilation, it labels the sure regions as foreground and background. The regions which are not sure are labeled as zero. Then, it updates the labels as segmentation procedure proceeds. It is an interactive segmentation process. It was used to count the number of buildings and segment out rooftop locations from the map. Due to bad quality of Google-maps India, distance transform was not quite successful in separating rooftop from the background image. The algorithm was giving many false positives as depicted in Fig. \ref{fig3}.

\vspace{3mm}

\subsubsection{Gabor Filter}
It analyzes the region of interest and detects whether there is any specific frequency content in a particular direction is present or not.\cite{ref_6} This procedure of segmentation is based on Gaussian distribution technique. Gaussian Mixture Models(GMM) assumes all data points are a combination of Gaussian distribution functions. Gabor filter enhances one region relative to another depending on the frequency and theta values. The general mathematical equation for Gabor filter in two-dimensional form is as follows:
\begin{equation}
    \textit{G(x,y; f,theta)} = exp(-1/2[\frac{x^2}{(delta_x)^2}+ \frac{y^2}{(delta_y)^2}])cos(pi*f*x)
\end{equation}

\begin{figure}[htbp]
\centering
\includegraphics[width=0.8\linewidth]{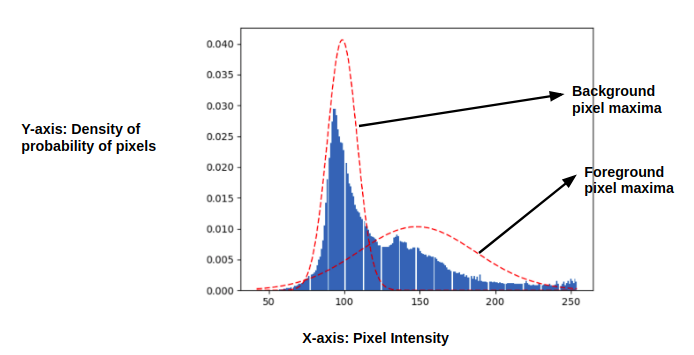}
\caption{Two Gaussian Mixture models were fixed over the satellite image to separate the foreground and background.} \label{fig4}
\end{figure}

After applying this filter, two GMMs were fit in the image histogram to separate out foreground and background. The output after applying two models is shown in Fig. \ref{fig4}.

\vspace{3mm}    

\subsubsection{Active Contours}
This algorithm is also known as \textit{snakes algorithm} are programmed to generate curves that move within images to find object boundaries. It is used to find object boundaries within an image. \cite{ref_7} \cite{ref_8} \cite{ref_9}  It is an iterative flow of gradient vector from the point of initiation in all directions. Active contours, the gradient vector flow (GVF), starts with the calculation of a field of forces, termed as GVF forces, over the image domain. The GVF forces drives the snake, represented as a physical object that has a resistance to both stretching and bending on the boundaries of the object. The generalized diffusion equations is applied to both components of the gradient of an image edge map to calculate the GVF forces.

\begin{figure}[htbp]
\centering
\includegraphics[scale=0.5]{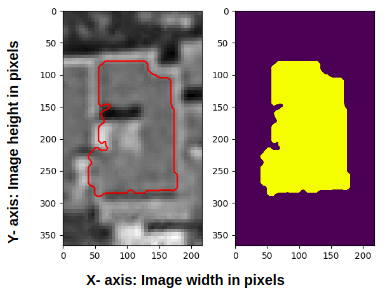}
\caption{Active Contours rooftop segmentation.} \label{fig5}
\end{figure}

As GVF forces are derived from a diffusion operation, they extend very far away from the object. This extends the "capture range" so that snakes can find objects that are very far away from the snake's initial position. This same diffusion creates forces that can pull active contours into concave regions. Hence, in our case it successfully detects the obstacles present inside the roof as well as mark the boundaries too. The result of Active Contour applied on edge sharpened image is shown in Fig. \ref{fig5}.

\vspace{3mm}
\subsubsection{Adaptive Canny Edge Detection}
\begin{figure*}[htbp]
\centering
\includegraphics[width=0.7\textwidth]{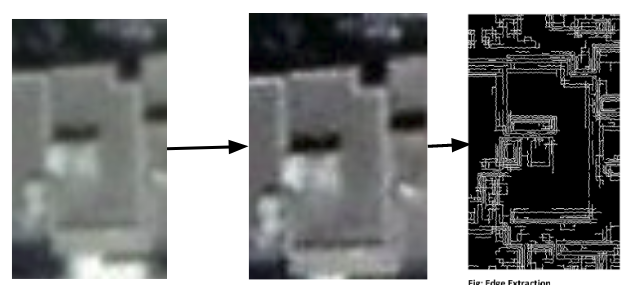}
\caption{Adaptive Canny Edge Detection results. The leftmost image is the original image. The center image is the bilateral sharpened image and the rightmost image is the output of the Edge Detection algorithm.} \label{fig6}
\end{figure*}

Based on the comparative analysis of edge detection algorithms, Canny edge detection algorithm outperforms others. Now, adaptive canny edge algorithm adapts itself based on the image.\cite{ref_10} The hysteresis threshold, in this case, is adaptive relevant to the image. The upper and lower threshold is calculated by using mean and variance of an image intensity. To enhance the edges in the low quality image, we applied Bilateral filter. A bilateral filter\cite{ref_23} is an edge-preserving filter which helps to reduce noise while maintaining the smoothness of the images. It is a non-linear filter. The output of the Canny Edge Detection algorithm also contains many false positives and it was also very haphazard as shown in Fig.\ref{fig6}.

\subsection{Polygon Shape Approximation}
The shape of the polygon that we get from Active Contours as output is very much distorted. We devised the following procedure to obtain a perfect polygon.

\subsubsection{Hough Transform} It is used to detect any shape that can be represented in a mathematical form.\cite{ref_11} and \cite{ref_12} was used to analyze the shape of a rooftop. Using K-Means clustering, the number of Hough line Transform were reduced to 4 to 6 to outline the rooftop shape alongwith objects present inside the rooftop.

\subsubsection{Pixel-based Polygon filling} Applying contour detection algorithm on the shape we got from Active contour, we moved clockwise around the contour pixel-wise. Each pixel on the contour and its surrounding pixels were marked as rooftop area. Despite this, we got a distorted polygon shape. Then, we moved on to region-based Polygon filling.

\vspace{3mm}

\subsubsection{Region-based Polygon filling} 
\begin{figure}[htbp]
\centering
\includegraphics[width=0.9\linewidth]{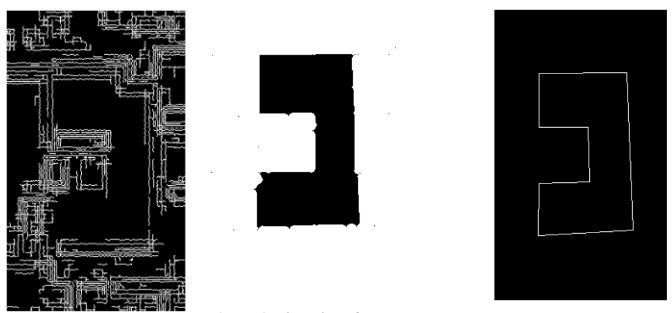}
\caption{Polygon Shape Approximation on Canny Edge detection output using K-means Clustering Algorithm. The first image is the Canny Edge map image. The second image shows approximate polygon after applying Region-based polygon filling. Rightmost image shows the final polygon shape after further refinement.} \label{fig7}
\end{figure}

After applying Hough Transform in combination with K-Means clustering, the rooftop area was divided into different regions. Checking the intensity of different patches, the area was marked as a rooftop area or not. If the mean intensity is greater than the threshold value, then that region is marked as a rooftop area and it is filled with black color on a different white image. The approximate shape of a rooftop after applying Region-based filling is shown in Fig.\ref{fig7}.

\section{Experiments and Results Analysis}
\subsection{Edge Detection Algorithms}
We evaluated various types of edge detection algorithms to detect rooftop boundaries.\cite{ref_14} \cite{ref_15} \cite{ref_16} Adaptive Canny outperforms other algorithms such as Laplacian of Gaussian, Sobel, Robert and Prewit. The parameters for all the other edge detection algorithms has to be set manually and thus can't be generalized. Adaptive Canny automatically analyses the image, hence, producing the best edge detection results for each image.


\begin{figure*}[htbp]
\centering
\includegraphics[width=0.8\textwidth]{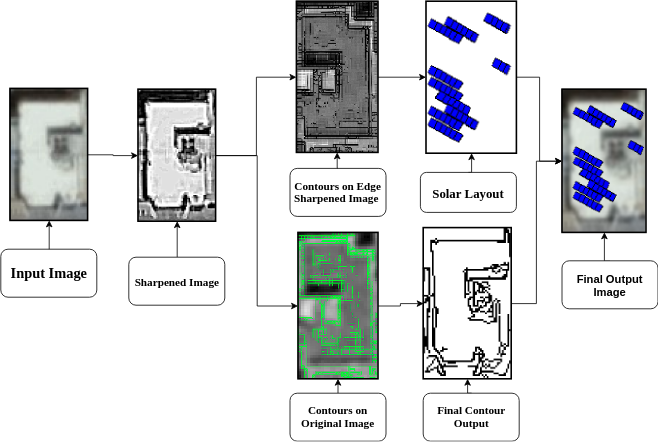}
\caption{Proposed Architecture of our Solar Panel Placement System} \label{fig9}
\end{figure*}

\subsection{Optimal Rooftop Area}

Using single feature characteristic of an image was not helpful to carve out the optimal rooftop area. It always leads to many false positives, and, was not able to detect obstacles present inside the rooftops too. Henceforth, we employed two features at the same time to get the exact rooftop area. The two methods which we used to get the shape of the rooftop is as follows:

\subsubsection{Harris Corners \& Adaptive Canny} 
Where Corner features and Canny Edge results were overlapping, those points were selected. The problem with corner features are that they come in very haphazard manner. They can't be accesses clockwise or anti-clockwise to estimate a Polygon shape. Therefore, to find the optimal rooftop area was impossible to find through this method. As Harris corners were of no use, so we discarded the possibility of Harris Corner features and Contour features overlapping case.

\vspace{3mm}

\begin{figure*}[htbp]
\centering
\includegraphics[width=0.8\textwidth]{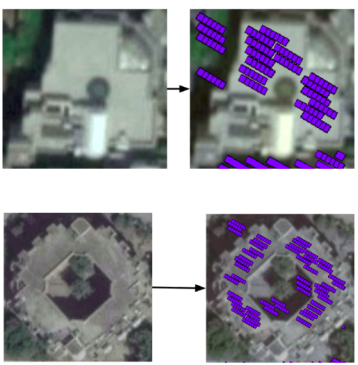}
\caption{Proposed Architecture output. The input image is on the left side and the blue solar panels area shows the optimal area available for solar panel placement with proper angle orientation. In the image, we can see in the first case, there are some solar panels plotted outside the rooftop area.} \label{fig8}
\end{figure*}

\subsubsection{Adaptive Canny \& Contours}
Contours can be accessed in a clockwise manner. When both these features are used simultaneously, then we got the most favorable number of features that can be used to place Solar panels. On two images, Canny was applied. One is a bilateral edge sharpened image and the other is an adaptive canny edge map image. In Edge sharpened image, the contours with thresholding effect mark out the rooftop boundaries. Contours on the canny edge map using threshold gives obstacle boundaries on the rooftop. When we take union over both results and plot it on a new white patch, it gives the most exact rooftop optimal area for solar panels placement compared to our all previous methods. Two thresholds were used one is the area of each contour and another one is a number of points used to combine the contour. At last, we got the optimal rooftop area for solar panel placement. The complete architecture of our work flow is shown in Fig.\ref{fig9}.

\vspace{3mm}

\subsubsection{Solar Panel Placement} We considered solar panels patches of 5x1, 4x1, 3x1. We also took into account the orientation angle required for maximum solar power output at a certain longitude and latitude. After placing solar panels horizontally, the main challenge was to rotate the panels with a certain angle, because, it will go out of the rooftop. For rotation of patches, I took a region of interest then rotated it by the amount of angle required. Then, we iterated along the breadth and length and evaluated the mean intensity of the region of interest using line iterator. If the mean intensity is 255, then the area is available for solar panel placement. The experimental results on the rooftops dataset are shown in the Fig. \ref{fig8}. 

\section{Conclusion and Future Work}

The paper presented several approaches on which thorough analysis was done to segment out optimal rooftop area and placement of solar panels too. In India, due to the low quality of satellite images, it is tough to analyze any location whether an area has solar potential or not. Many people are also unaware of how much solar power their rooftop can produce. Our solution can provide a review of a rooftop remotely, and it motivates the people to use more and more renewable energy somewhat totally dependent on the non-renewable sources of energy. We focused on the Adaptive Edge Detection and Contours to segment out rooftop boundaries and obstacles present inside them accurately. Our approach targets the installation of solar panels and provides a comparative analysis of the solar potential of a building or a house. Previous works done on solar panels placement was done mainly in foreign countries where they have 3D mapping. 

As we can see in Fig. \ref{fig8}, there are some outliers that are plotting solar panels outside the building rooftop area. In the future, the author aims to improve the approach by using Deep learning approaches to learn substantial features. We are currently increasing our dataset for supervised classification of rooftop detection from Satellite images \cite{ref_18} using Mask R-CNN. \cite{ref_17} We are also considering rendering 3D depth reconstruction from a single image to improvise our analysis.\cite{ref_19} For improvement of edge detection and segmentation, we are currently exploring Gradient-based Edge-Aware filters \cite{ref_20} and Deep Active Contours\cite{ref_21} respectively.

{
\bibliographystyle{ieee}
\bibliography{ref}
}

\end{document}